\documentclass[letterpaper, 10pt, conference]{ieeeconf}  

\IEEEoverridecommandlockouts 

\usepackage{amsmath,amsfonts,amssymb,mathtools} 
\usepackage{bbm,bm,nicefrac} 
\usepackage{graphicx,float} 
	\graphicspath{{./images/}}
\usepackage[table,xcdraw]{xcolor}
\usepackage{algorithm,algpseudocode} 
\let\labelindent\relax \usepackage[inline]{enumitem}
\usepackage{cite}
\usepackage{cases}
\usepackage{bm}
\usepackage{soul,color}

\newtheorem{remark}{Remark}

\newtheorem{problem}{Problem}

\algnewcommand{\algorithmicand}{\textbf{ and }}
\algnewcommand{\algorithmicor}{\textbf{ or }}
\algnewcommand{\OR}{\algorithmicor}
\algnewcommand{\AND}{\algorithmicand}
\algnewcommand{\var}{\texttt}

\begin{document}

\title{\LARGE \bf Don't Get Stuck: A Deadlock Recovery Approach\\

\author{Francesca Baldini$^{1}$, Faizan M. Tariq$^{1}$, Sangjae Bae$^{1}$ and David Isele$^{1}$
\thanks{$^{1}$Honda Research Institute USA, San Jose, CA 95134, USA.
        Email: {\tt\small \{francesca\_baldini, faizan\_tariq, sbae, disele\}@honda-ri.com}
}%
}
}
\maketitle

\begin{abstract}

When multiple agents share space, interactions can lead to deadlocks, where no agent can advance towards its goal. This paper addresses this challenge with a deadlock recovery strategy. In particular, the proposed algorithm integrates hybrid-A$^\star$, STL, and MPPI frameworks. Specifically, hybrid-A$^\star$ generates a reference path, STL defines a goal (deadlock avoidance) and associated constraints (w.r.t. traffic rules), and MPPI refines the path and speed accordingly. This STL-MPPI framework ensures system compliance to specifications and dynamics while ensuring the safety of the resulting maneuvers, indicating a strong potential for application to complex traffic scenarios (and rules) in practice. Validation studies are conducted in simulations and on scaled cars, respectively, to demonstrate the effectiveness of the proposed algorithm.


\end{abstract}



\section{Introduction}

\begin{figure}
    \centering
        \includegraphics[width=1\columnwidth]{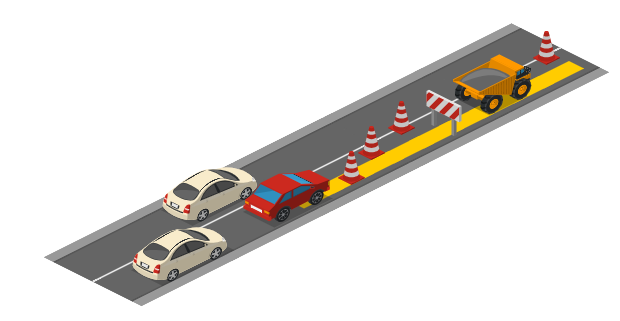}
        \caption{In the depicted scenario, the ego vehicle, marked in red, faces a deadlock within a construction zone while following its predefined global path, indicated by a yellow line. To resolve the deadlock, the vehicle needs to create a new path that circumvents the construction area, avoids collisions with other vehicles, and allows it to continue on its intended route.}
        \label{fig:motivation}

\end{figure}

Deadlocks, such as cars heading in opposite directions through a narrow passage, produce challenging problems that can be difficult for human drivers and autonomous vehicles (AVs) to resolve (as illustrated in Fig.~\ref{fig:motivation} and Fig.~\ref{fig:image2}).  These situations, which require intricate agent prediction,  routing and rerouting strategies, and navigation through expanded dynamic spaces, make resolving deadlocks a complex issue for AV technology. 

Since the challenges of deadlocks were recognized \cite{de2015autonomous}, research activities have been initiated to reduce or prevent deadlocks for autonomous vehicles~\cite{Raphael,moller2023model}. However, it is often not possible to avoid or prevent deadlock situations -- as that also depends on other traffic agents. Thus, ``recovery'' strategies from the deadlock scenarios are necessitated as part of fully automated driving. Real-world issues faced by dispatched level-4 automated cars (e.g., Waymo and Cruise) further support the motivation of this study, where the automated cars are stuck at construction sites \cite{waymoStuck}, obstacles, and/or other cars \cite{cruiseStuck}. This paper contributes to the literature (as well as the facilitation of the deployment of automated cars) by introducing an explainable, mathematically rigorous, and computationally efficient deadlock recovery scheme.

Despite its importance, the deadlock recovery strategy has been relatively unexplored. Relevant studies have focused particularly on solutions with connectivity and cooperation of other cars \cite{adrive2022,perronnet2019deadlock,deadlockdet,Goto}. However, a deadlock recovery strategy for a single agent in \textit{the absence of connectivity and cooperation} has not been thoroughly explored (if not entirely), and this study fills the research gap by focusing on a single-agent deadlock recovery. 


We define ``deadlock'' as any condition where our vehicle, due to conflicting goals or physical obstructions, finds itself unable to move forward. This presents concerns for traffic flow and safety, especially in urban settings where real-time decision-making is crucial and where AVs must coexist with human-driven vehicles, each relying on distinct decision-making processes. Note that in this work, we focus on the simpler case of static obstacles, and therefore consider not only cases of mutual deadlock, but also cases where the other agents \emph{could} move to create forward progress, but do not.   
While human-driven vehicles navigate such situations through 
human judgment and gesturing, AVs require a more systematic approach.


Our research 
establishes a framework that permits the real-time regeneration of vehicle paths while maintaining safety in all maneuvers. Building upon the foundations laid by \cite{hybridastar}, our approach creates as a first step non-holonomic paths that adhere to the dynamic capabilities of the AV and comply with spatial and environmental constraints. To ensure safe execution of operations to break free from a deadlock, we develop a strategy to combine STL predicates ~\cite{stl} with the Model Predictive Path Integral controller\cite{mppi}. This integration leverages stochastic sampling to significantly reduce the computational demands typically associated with classical control methods such as MPC, enhancing safety and compliance with traffic laws during deadlock recovery.

Importantly, our validation studies include both simulations and physical system experiments (with 1/10 scale race cars). Simulation studies allow us to conduct an in-depth analysis and validation of the proposed method under controlled conditions. Alternatively, the physical system experiments showcase the proposed approach's practical applicability (while examining the sim-to-real gap).


\section{Background and Related works}

\subsection{Deadlocks}
Deadlock situations in autonomous driving present unique challenges that extend beyond the scope of traditional path and motion planners. These scenarios often require the vehicle to execute reverse maneuvers and utilize extreme steering angles, which are not typically prioritized in standard driving algorithms \cite{slas, bidirectionalOvertaking}. Furthermore, resolving deadlocks may require compromising passenger comfort to navigate out of tight spaces, emphasizing maneuverability over smoothness \cite{rcms}. This distinct set of requirements highlights the need for specialized planning techniques that can dynamically adapt to such demanding driving conditions.

Traffic deadlocks - situations where vehicles obstruct each other and halt movement - greatly impact urban traffic flow and safety. Although there has been notable progress, a large portion of research on solving deadlocks assumes that vehicles can communicate with each other through vehicle-to-vehicle (V2V) technology. This assumption narrows the scope of these solutions, as not all vehicles are currently equipped with or capable of using V2V communication effectively \cite{cooperativeOvertaking}.

Recent advances in autonomous vehicle navigation have focused on addressing the complexities of urban traffic, particularly deadlock detection and recovery. A significant contribution to this field is the A-DRIVE system\cite{adrive2022}, which provides a comprehensive solution for the detection and recovery of deadlocks at road intersections for connected and automated vehicles. 
Despite its effectiveness, A-DRIVE's deployment relies heavily on advanced V2V communications.

The study referenced in \cite{deadlockdet} addresses the gaps in current research, which often relies on predetermined paths or assumes that vehicles operate within well-defined road layouts, by expanding the investigation to account for the dynamics of human-driven vehicles (HDVs) in less organized settings. It introduces a classification of deadlocks into two categories (weak and strong) and presents two novel detection algorithms based on the concept of propagation of evasion distances. 
 
 The work by \cite{wang} introduces the concept of Altruistic Cooperative Driving (ACD), a novel strategy that leverages vehicle-to-vehicle (V2V) communication to enable CAVs to share information and driving intentions with their counterparts. 
Similarly, \cite{perronnet} underscores the feasibility of achieving deadlock-free traffic flow through a novel graph-based model that effectively captures the dynamics of intersection networks, by introducing a hierarchical framework that leverages the capabilities of network servers, intersection servers, and the vehicles themselves.

The work from \cite{Goto} introduces a pioneering approach utilizing multi-agent deep reinforcement learning to enhance path planning capabilities, 
However, the requirement for extensive training data to accurately model diverse and dynamic traffic scenarios may limit the immediate applicability of this approach.

\begin{figure}[ht!]
    \centering
        \includegraphics[width=\linewidth]{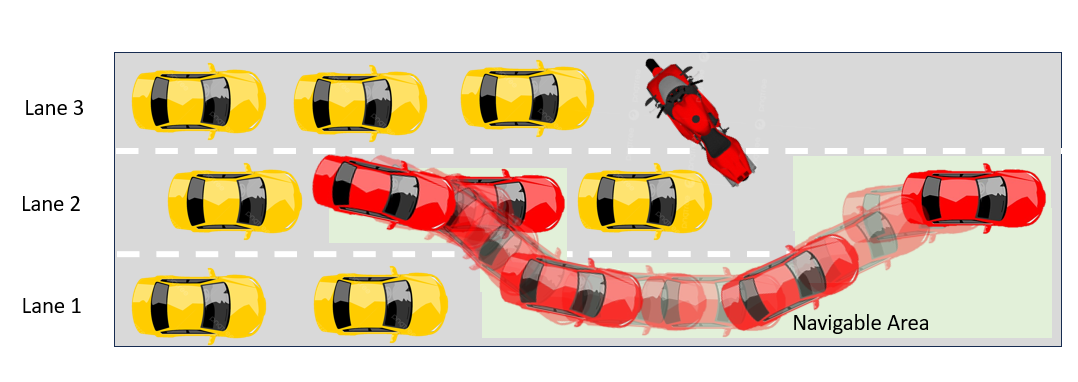}
        \caption{In this scenario, the ego vehicle confronts a deadlock caused by a motorcycle crash. To escape, it must perform complex backward and forward maneuvers to safely navigate around the crash site and continue its journey without colliding with other vehicles. }
        \label{fig:image2}
   
\end{figure}

\subsection{Motion Planning under Signal Temporal Logic Constraints}
Temporal logic offers a mathematical structure for defining the intended behaviors of a system. Signal Temporal Logic (STL) was introduced to outline and oversee the anticipated performance of physical systems. A key benefit of STL is its ability to quantitatively assess how well the specified properties are met or violated. STL has emerged as a powerful tool in motion planning for its precision in specifying temporal and spatial requirements for robotic systems. 

Fusing STL specifications with control theoretic approaches, such as MPC \cite{murray, stlcontrol1}, has led to sophisticated and reliable motion planning frameworks. This integration enables the automatic generation of control policies that guarantee adherence to STL specifications, even in complex and unpredictable environments. To achieve this, the dynamics and the STL specifications are encoded as constraints of the optimization problem.
However, this presents several challenges and limitations. Specifically, MPC requires solving an optimization problem at each control step, considering the predicted behavior of the system over a horizon. Incorporating STL constraints, which can be highly complex and non-linear, increases the computational cost significantly. This can limit the real-time applicability of the approach, especially for systems that require fast decision-making in dynamic environments.

The work proposed by \cite{rulebasedcontrol} presents an approach to maintaining safety and regulatory compliance by integrating Control Barrier Functions (CBFs) and Control Lyapunov Functions (CLFs) to ensure system stability while adhering to safety constraints. It introduces rulebooks based on traffic laws and cultural norms to guide the behavior of autonomous vehicles. These rulebooks, incorporating Signal Temporal Logic (STL) for evaluating vehicle trajectories, aim to ensure compliance with rules at all times. However, the complexity of accurately quantifying rule violations and the variability in traffic conditions can pose challenges in practical application. Other research in this field~\cite{kapoor, sahin, Cho} focuses additionally on the use of STL  to formalize traffic rules and safety constraints for autonomous vehicles, ensuring that they operate within the bounds of these rules while navigating complex traffic environments. 
Finally, Halder et al.~\cite{halder} developed an A$^\star$-based velocity planner for autonomous shuttles that prioritizes constraints hierarchically, ensuring that more critical constraints are violated last when not all can be simultaneously met. This approach offers a practical solution to handling unpredictable behaviors in dynamic traffic environments. 

\subsection{Motivation}
STL is a formalism that is used to specify properties of trajectories in a way that allows for the precise definition of temporal and logical constraints. In the context of autonomous driving, STL can be employed to describe desired behaviors of the vehicle over time, such as safety margins, speed limits, and temporal constraints on actions.
A Model Predictive Path Integral (MPPI) controller is an optimization-based control strategy that uses stochastic sampling to generate a distribution of possible future trajectories, evaluates them based on a cost function, and selects the optimal path to follow. This approach is useful in dynamic and uncertain environments, making it suitable for handling complex scenarios like deadlock recovery.
Combining STL specifications with MPPI control can provide a robust framework for deadlock recovery in autonomous driving by ensuring that the vehicle not only follows the optimal path to recover from deadlock situations but also adheres to the specified safety and temporal constraints. 

\subsection{Contribution}

Our work presents an advancement in resolving deadlock situations for automated driving. By integrating STL specifications into the MPPI control framework, 
we enhance the vehicle’s ability to navigate complex traffic scenarios. The key contributions of this research 
is the proposal of a hybrid approach that leverages the robustness of MPPI in handling uncertainties with the precise constraint enforcement capabilities of STL. This dual framework allows for real-time adherence to dynamical safety and operational constraints.
    
We use STL to formally define and enforce safety margins, speed limits, and other critical temporal constraints within the control framework. This ensures that all vehicle maneuvers adhere to predefined safety standards even under uncertain conditions and narrowed areas. 
    

\section{Problem Statement}
We formally define our deadlock recovery problem as:
\begin{problem}\label{problem}
Given the initial state of the ego vehicle $\mathbf{x}^{ego}_{0}$, its dynamical constraints, the current state of other vehicles $\mathbf{x}^i_{curr}$ , where $i \in \mathbb{N}$,  
and a set of traffic rules $\Phi$,  find a sequence of control inputs $\mathbf{u}_t = \{u_1, u_2, \ldots, u_{T-1}\}$ 
to transition the vehicle from $\mathbf{x}^{ego}_{0}$ to the goal position $\mathbf{x}^{ego}_T$ for recovering from the deadlock while avoiding collisions with the surroundings vehicles, and complying with traffic laws.
\end{problem}

Fig.~\ref{fig:image2} illustrates a targeting scenario for deadlock recovery. To solve the problem, it requires: (i) detection of deadlock conditions, and (ii) list of traffic laws to adhere and (safety) constraints to adhere.

  

\subsection{Deadlock Conditions}

Aligned with the definition earlier, the ego vehicle is said to be in a deadlock situation when it is unable to continue on its intended path or reaching its destination. 
Deadlock scenarios may arise from situations where multiple agents, including vehicles and pedestrians, obstruct each other's progress, creating a standstill where no participant can advance without the cooperation or movement of others.
In this study, we specifically tackle deadlock situations 
assuming that surrounding obstacles and vehicles remain stationary. 
 
 \begin{remark}
 In some deadlock scenarios, it may require to evaluate other agents' behaviors reactive to the ego vehicle's actions. Evaluating inter-agent interactions for deadlock recovery remains for future work.    
 \end{remark}

\begin{table*}[ht]
    \centering
    \begin{tabular}{|c|c|}
    \hline
      \textbf{ Description}  & \textbf{Rule}  \\
     \hline
    \hline

      Keep safety distance   & $\mathbf{G} (\text{dist}(t) \geq d_{safe}) $ \\
     \hline

    Avoidance Collision   
    & $\mathbf{G} \left( \bigwedge_{i} \left( x(t) < x_{\text{min, i}} \lor x(t) > x_{\text{max, i}} \lor y(t) < y_{\text{min, i}} \lor y(t) > y_{\text{max, i}} \right) \right)$
 \\
     \hline

      Stay in the lane   &  $\mathbf{G} \left( \bigwedge_{i} \left( x_{\text{min, i}} \leq x(t) \leq x_{\text{max, i}} \land y_{\text{min, i}} \leq y(t) \leq y_{\text{max, i}} \right) \right)
$ \\
     \hline
    Do not go on the oncoming lane until the oncoming lane path is free & $\mathbf{G} \left( \neg \text{IncomingLane}(t) \, \mathbf{U} \, \text{isFree}(t) \right)
$\\
    \hline
      Do not cross the intersection until first stopped    & $\mathbf{G} \left( \neg \text{crossing}(t) \, \mathbf{U} \, \left( \text{stopped}(t) \land \mathbf{F}_{[0,3]} \, \text{stopped}(t + 3) \right) \right)$ \\
     \hline

      

     Eventually, exit the deadlock & $\mathbf{F} \, (\text{exitDeadlock}(t))
$\\
\hline
     
    Minimize abrupt changes in velocity & $\mathbf{G} \left( \text{changeInVelocity}(t) \leq \Delta v_{\text{max}} \right)
$\\
    \hline
    
    \end{tabular}
    \caption{STL Rules considered in our scenarios. More rules can be incorporated as we extend to more dynamic environments. }
    \label{tab:rules}
\end{table*}

\section{Proposed Approach}   \label{sec:stl-mppi}
\begin{figure}
    \centering
    \includegraphics[width=0.7\columnwidth]{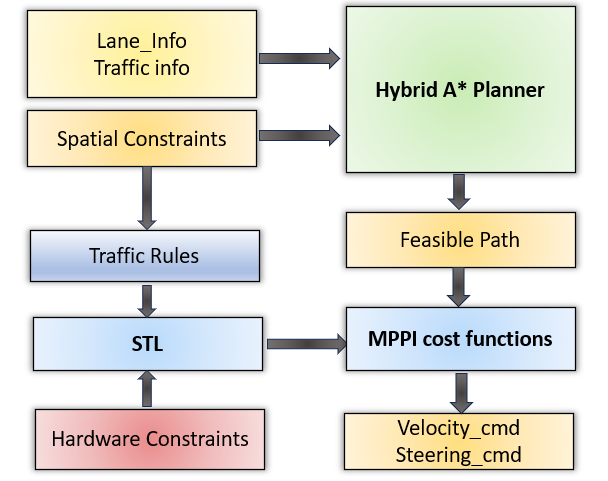}
    \caption{The proposed system integrates traffic, lane, and spatial data into a Hybrid A$^\star$ planner to formulate a path for an autonomous vehicle. This path informs the MPPI controller, which integrates traffic rules and vehicle constraints as STL specifications. The controller then outputs commands that align with these rules while maintaining proximity to the planned path.}
    \label{fig:pipeline}
\end{figure}

Fig.~\ref{fig:pipeline} illustrates the overall pipeline of the STL-MPPI framework.
Overall, STL-MPPI refines control inputs that minimize a cost function. The cost function balances the goal achievement (i.e., recovery) and adherence to safety and constraints governed by STL. 

\begin{figure}
    \centering
    \begin{minipage}{\linewidth}
        \centering
        \includegraphics[width=0.49\linewidth]{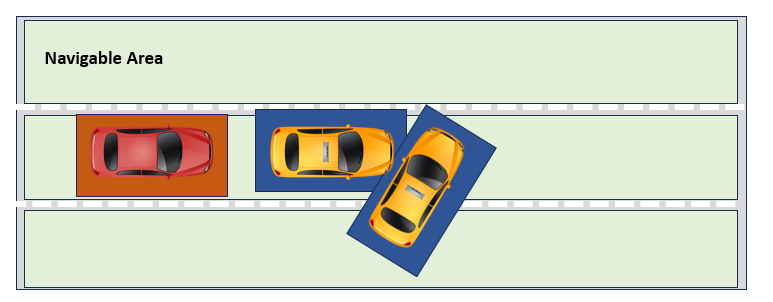}
        \includegraphics[width=0.49\linewidth]{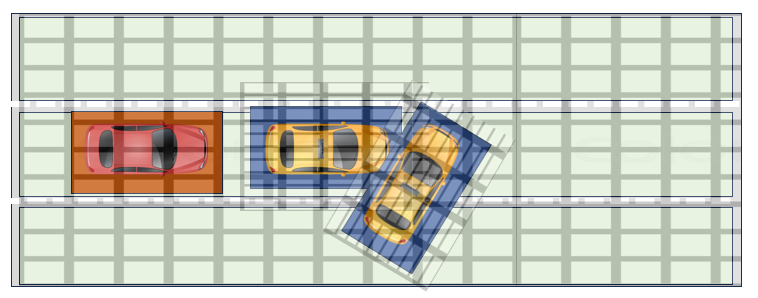}
        \caption{The navigable space in the system is discretized into sparse grids along the lanes and finer grids around obstacles. This approach ensures that even in tighter spaces, a feasible path is always available.}
        \label{fig:h1}
    \end{minipage}
    \hfill  
    \begin{minipage}{\linewidth}
       \vspace{0.5cm}
        \centering
        \includegraphics[width=0.49\linewidth]{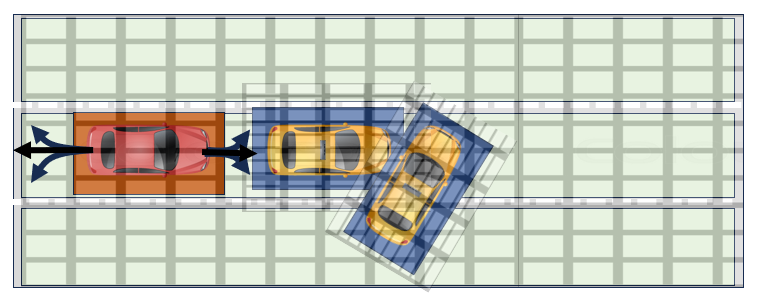}
                \includegraphics[width=0.49\linewidth]{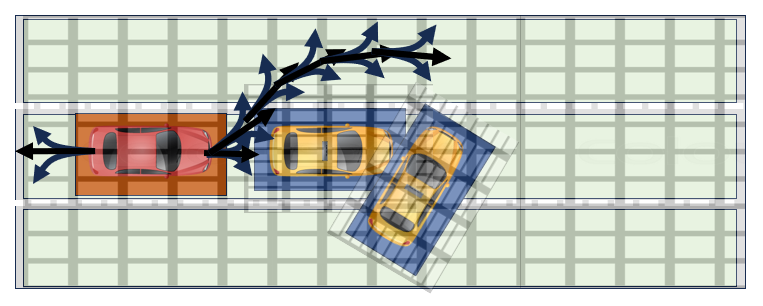}
        \caption{The Hybrid A$^\star$ algorithm generates a feasible path for vehicles by adhering to vehicle dynamics and spatial constraints, such as staying within lane boundaries and maintaining safe distances from obstacles.}
        \label{fig:h2}
        \hfill
    \end{minipage}
    \hfill  
    \begin{minipage}{\linewidth}
       \vspace{0.5cm}
        \centering
        \includegraphics[width=0.49\linewidth]{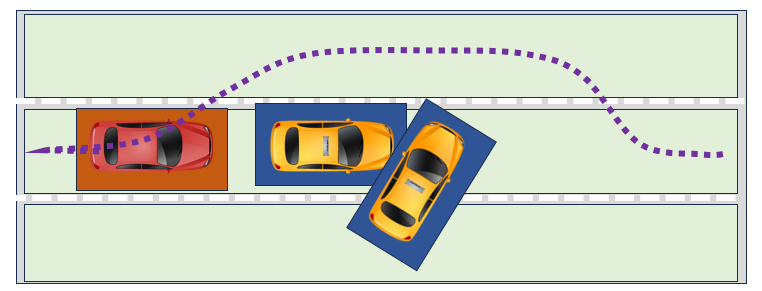}
                \includegraphics[width=0.49\linewidth]{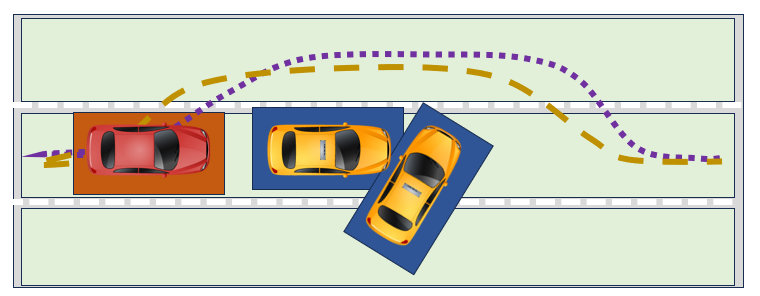}
        \caption{After the Hybrid A$^\star$ algorithm creates an initial path (purple dashed line), the STL-MPPI controller dynamically generates a velocity and steering profile that results in a new path (yellow dashed line), aligning with spatiotemporal constraints and hardware specifications.}
        \label{fig:h3}
        
    \end{minipage}
    
\end{figure}

The process begins with a modified version of the Hybrid $A^\star$ planner that crafts a feasible initial path $\tau$, considering both vehicle dynamics and environmental constraints as illustrated in Fig.~\ref{fig:h1} and Fig.~\ref{fig:h2}. This path is then refined through the STL-MPPI method (Fig.~\ref{fig:h3}).

\paragraph{Signal Temporal Logic (STL)}
We employ STL to adhere to traffic rules and constraints for the ego vehicle, specifying temporal properties of a desired trajectory. For a signal $\mathbf{x}_t$, where $t \in \mathbb{R}_{\geq 0}$ represents time, an STL formula $\phi$ can be defined recursively as follows:
\begin{equation}
    \Phi ::= \top \,|\, f(x) > 0 \,|\, \neg \phi \,|\, \phi \wedge \psi \,|\, F_{[a,b]} \phi \,|\, G_{[a,b]} \phi \,|\, \phi U_{[a,b]} \psi
\end{equation} 
    where:
    
    \begin{itemize}
        \item $\top$ signifies a condition that is always true.
        \item $f(x) > 0$ denotes an atomic proposition defined by a function ff of the signal $\mathbf{x}_t$.
        \item $\neg$, $\wedge$, $F$, $G$, and $U$ represent negation, conjunction, future, globally, and until operators, respectively.
        \item $a, b \in \mathbb{R}_{\geq 0}$, with $a \leq b$, specify the time interval for temporal operators.
    \end{itemize}
  

STL specifications $\Phi$ for deadlock recovery encompass safety constraints, such as maintaining a safe distance from other vehicles $G_{[0,T]} (d(\mathbf{x}_t) \geq d_{safe})$, where $d(\mathbf{x}_t)$ denotes the distance to the nearest vehicle and $d_{safe}$ represents a predetermined safe threshold distance. We additionally consider temporal properties, such as stopping at intersections as mandated by road rules $(F_{[0,T]} \textit{stop at intersection})$, and liveness properties, like eventually leaving the deadlock zone $(F_{[0,T]} \textit{outside deadlock zone})$. Table~\ref{tab:rules} reports the rules considered in our work. These rules can be extended for more dynamic environments.

\paragraph{STL-MPPI}

To integrate STL into the MPPI framework, we introduce penalty functions for violations of STL criteria within the cost function. These penalties $\mathcal{P}_\phi(\mathbf{x}_t, t)$ escalate the cost for any breach of STL specifications, steering the optimization process toward solutions that align with predetermined behavioral norms.

The MPPI controller optimizes control inputs over a finite horizon to minimize a cost function, which incorporates the trajectory's performance and compliance with STL specifications. The optimization problem is formulated as:
\begin{equation}
\min_{\{\mathbf{u}_t\}_{t=0}^{T-1}} E\left[ \sum_{t=0}^{T-1} \left(L(\mathbf{x}_t, \mathbf{u}_t) + \sum_{\phi \in \Phi} W_\phi \mathbf{I}_{\phi}(\mathbf{x}_t) \right)+ F(\mathbf{x}_T)\right]
\end{equation}
where:
\begin{itemize}
    \item $\mathbf{x}_t$ and $\mathbf{u}_t$ are the state and control input of the vehicle at time $t$, respectively.
    \item $T$ is the optimization time horizon.
    \item $L(\mathbf{x}_t, \mathbf{u}_t)$ is the running cost at time $t$, including terms for tracking a reference path.
    \item $F(\mathbf{x}_T)$ is the terminal cost, reflecting the desirability of the final state. 
    \item $\Phi$ is a set of STL specifications $\phi$ relevant to deadlock recovery.
    \item $W_\phi$ are weights reflecting the importance of satisfying each STL specification $\phi$.
    \item $\mathcal{P}_\phi(\mathbf{x}_t, t)$ is a penalty function for the violation of STL specification $\phi$ at time $t$, defined based on the robustness metric of STL.
\end{itemize}

The MPPI algorithm evaluates various control sequences $\mathbf{u}_k$, projecting the vehicle's path over the forecast horizon $T$.
For a trajectory $\tau_k$, the STL evaluation augments the cost function with penalties for violating STL constraints:
\begin{equation}
\mathbf{I}_{ \phi}(\tau_k) = 
\begin{cases}
0, & \text{if $\tau_k$ satisfies $\phi$},\\
\text{penalty}, & \text{otherwise}.
\end{cases}
\end{equation}

The optimal trajectory is chosen based on the computed costs. We evaluate and assign weights to trajectories $\tau_k$ according to their associated costs, giving preference to paths with lower costs, as determined by:
\begin{equation}
w_k = \frac{\exp\left(-\frac{1}{\lambda} J(\tau_k)\right)}{\sum_{j=1}^{K} \exp\left(-\frac{1}{\lambda} J(\tau_j)\right)},
\end{equation}
where $\lambda$ acts as the temperature parameter that regulates the distribution of weights among the trajectories. This mechanism selects the optimal trajectory that minimizes the cost function while meeting STL constraints, thereby ensuring safe and efficient navigation of the vehicle to its destination.

The sampling of control sequences $\mathbf{u}_k$ within the STL-MPPI algorithm is guided by a distribution centered around the current best estimate of the control sequence. The sampled control sequences are perturbed versions of this estimate, enabling the exploration of the control space to identify a sequence that minimizes the expected cost. The formula for generating these samples is as follows:
\begin{equation}
\mathbf{u}_k(t) = \mathbf{\bar{u}}_t + \epsilon_{k_t}, \quad \epsilon_{k_t} \sim \mathcal{N}(0, \Sigma),
\end{equation}
where:
\begin{itemize}
    \item $\mathbf{\bar{u}}_t$  represents the current best estimate of the control at time $t$.
    \item $\epsilon_{k_t} $ is a perturbation added to the control signal, sampled from a Gaussian distribution with mean 0 and covariance $\Sigma$.
\end{itemize}

This approach enables the MPPI algorithm to iteratively refine the control sequence by evaluating a diverse set of trajectories and selecting the one that offers the best balance between performance and compliance with the specified constraints.
By assigning appropriate weights to these STL-based constraints within the cost function, the controller prioritizes the satisfaction of these specifications. Higher weights on critical constraints ensure that the optimization process heavily penalizes any deviation from desired behaviors. This weighted approach allows the MPPI controller to balance the trade-off between minimizing cost and satisfying constraints, ensuring that essential specifications are met. Additionally, the receding horizon nature of MPPI, combined with real-time feedback, continuously updates and refines the trajectory, maintaining compliance with STL constraints and approaching optimality even in the presence of uncertainties and dynamic changes in the environment.

\section{Simulation}
We evaluate our proposed method on simulated driving scenarios. 
The vehicle dynamics are described by the 4D kinematic bicycle model in \cite{zang} discretized with a time step of $\Delta t = 0.1 s$. All simulations are performed in Python 3.8 on a Ubuntu 20.04 with an Intel Core i7-10700K CPU.
The parameter settings for the system are outlined as follows: The time horizon ($T_{hor}$) is set to 5 [sec], with a sampling count ($K_{samples}$) of 1000. Velocity variations are controlled by a standard deviation ($\sigma_{vel}$) of 0.12 [m/s], and steering variations by a standard deviation ($\sigma_{steer}$) of 0.35 [rad]. The trade-off parameter ($\lambda$) is set at 0.5 $\in[0,1]$. The weights for STL constraints and path adherence are respectively set to 1,000 ($\omega_{STL}$) and 100 ($\omega_{path}$). These parameters are crucial for managing the system's dynamics and ensuring adherence to predefined path and safety constraints.

 The task requires the ego vehicle to adjust its velocity to avoid collisions while following and changing lanes, and crossing intersections.
 We are considering an urban scenario and the perception systems give information about the lane boundaries and obstacles in both forward and oncoming lanes.
 The rules are defined from the perspective of the ego-vehicle.

Key parameters include maximum and minimum speeds, steering angle constraints, and vehicle and wheelbase dimensions.
The optimized path is simulated, showcasing the vehicle's navigation through the environment towards the goal region. Key metrics for evaluation include adherence to STL constraints, the efficiency of the navigated path, and the computational performance of the integrated Hybrid MPPI-STL approach. 

\subsection{MPC baseline}
To evaluate the performance improvements of our proposed MPPI-STL approach in terms of the key metrics outlined previously, we formulate an MPC-based reference tracking problem as:
\begin{align}
     \min_{\{\mathbf{x}_t, \mathbf{u}_t\}_{t=0}^{T-1}} \ & \sum_{t=0}^{T-1} L(\mathbf{x}_t, \mathbf{u}_t) + \delta \Phi(\mathbf{x}_T) + g(\mathbf{x}_t)\\
     \text{subject to:}&\\
     f(\mathbf{x}_t,\mathbf{u}_t) &=0,\forall t\in\{0,1,...,T\}\\
     \mathbf{x}_t &\in \mathcal{X},\forall t\in\{0,1,...,T\}\\
     \mathbf{u}_t &\in \mathcal{U}, \forall t\in\{0,1,...,T-1\},
\end{align}

where $L(\mathbf{x}_t, \mathbf{u}_t)$, $\mathbf{x}_t$, $\mathbf{u}_t$, $\delta$, and $\Phi(\mathbf{x}_T)$ are as defined in Section~\ref{sec:stl-mppi} while $g$ corresponds to the collision avoidance penalty given by a soft-max function \cite{boyd}, $f$ denotes the system dynamics given by the bicycle kinematic model \cite{zang}, $\mathcal{X}$ denotes the feasible set of $ \mathbf{x}_t$, corresponding to the drivable region limits, and $\mathcal{U}$ denotes the feasible set of $\mathbf{u}_t$, corresponding to the actuation (acceleration and steering) limits.

\begin {remark}
Hard safety constraints with ellipsoidal and rectangular obstacle models led to infeasibility issues or the vehicle staying stuck in a deadlock due to a conservative parametrization, all the while having a higher computational complexity. This made us switch to soft constraints; the penalty functions \cite{boyd} experimented with are the log-barrier, max, and soft-max functions.
\end{remark}

 \subsection{Simulations}


We compare the performance of our method against the MPC baseline in managing deadlocks under the following conditions:
\begin{itemize}
\item Deadlocks occurring within an intersection setting.
\item Deadlocks caused by roadway obstructions, such as accidents or road maintenance.
\item Deadlocks under conditions of dense and complex traffic patterns.
\end{itemize}










\paragraph{Intersection Deadlock}
In urban environments, particularly at intersections, a common occurrence is vehicles entering intersections without adequate space to exit, thereby obstructing the path for crossing traffic. This situation often leads to gridlock, significantly disrupting traffic flow. 

\begin{figure}[h!]
    \centering
    \begin{minipage}{\linewidth}
        \includegraphics[width=0.49\linewidth]{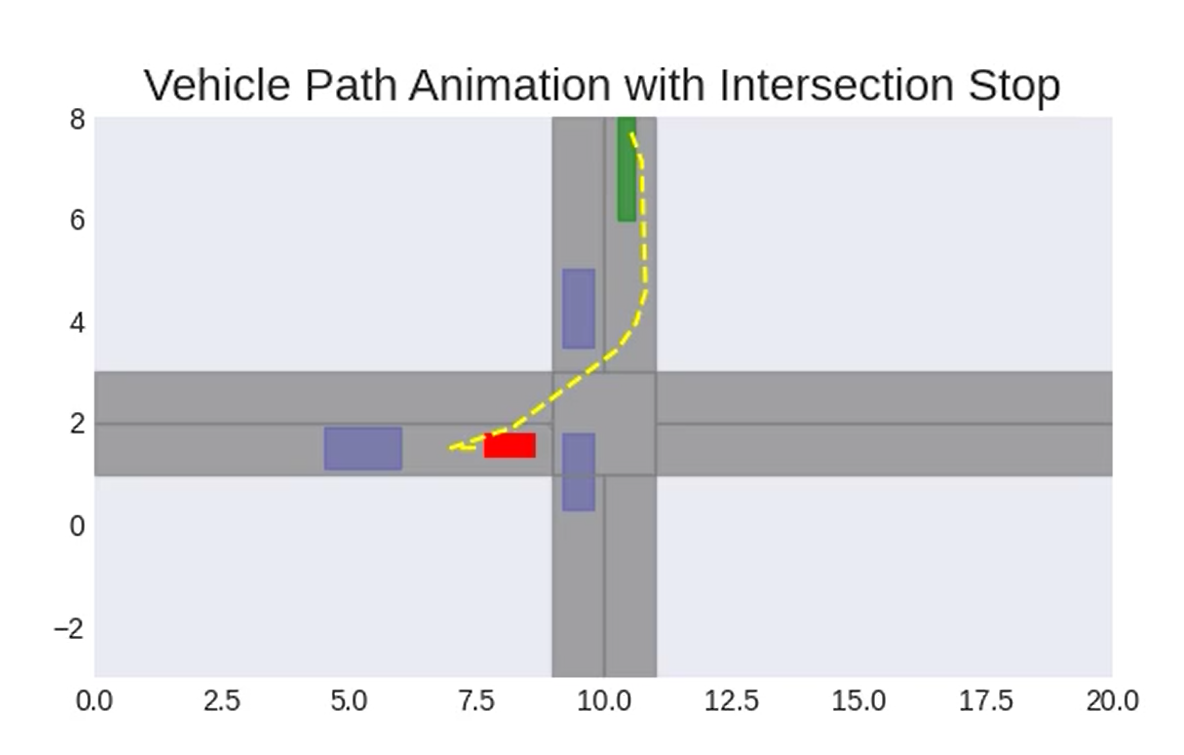}  
        \includegraphics[width=0.49\linewidth]{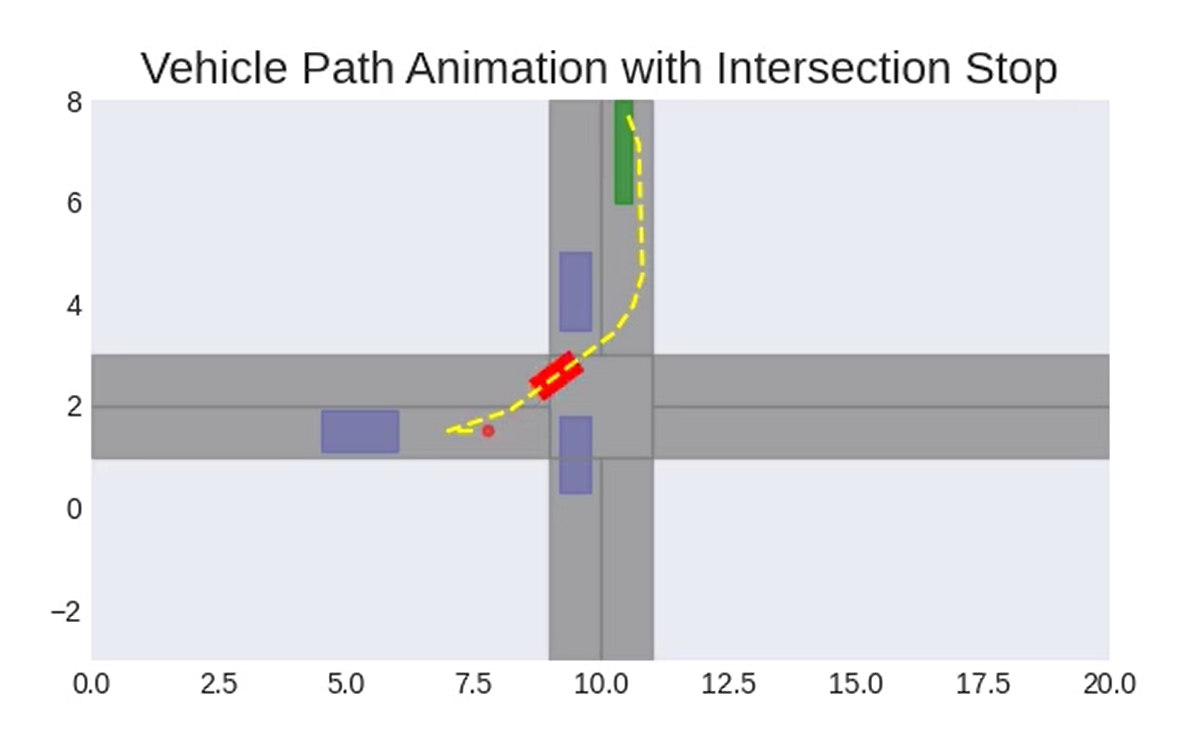}
        \caption{STL-MPPI path for intersection scenario}
        \label{fig:intersection}
    \end{minipage}
\end{figure}

Figure~\ref{fig:intersection} shows a scenario in which an autonomous vehicle is set to cross an intersection but finds its way blocked by another vehicle that has stopped, obstructing the AV's intended path.  
In this scenario, the AV must maintain a safe distance from the stopped cars and stay in its lane of incoming traffic. The AV must not leave its lane unnecessarily  and should stop at the intersection for 3 seconds before crossing, according to the road rules.
Our testing shows that while the MPC method could not navigate through an intersection deadlock scenario, the STL-MPPI successfully maneuvered the AV without collision and with negligible computational time (1.7e-5 seconds).

\begin{figure}[h!]
    \centering
    \begin{minipage}{\linewidth}
    \includegraphics[width=\linewidth]{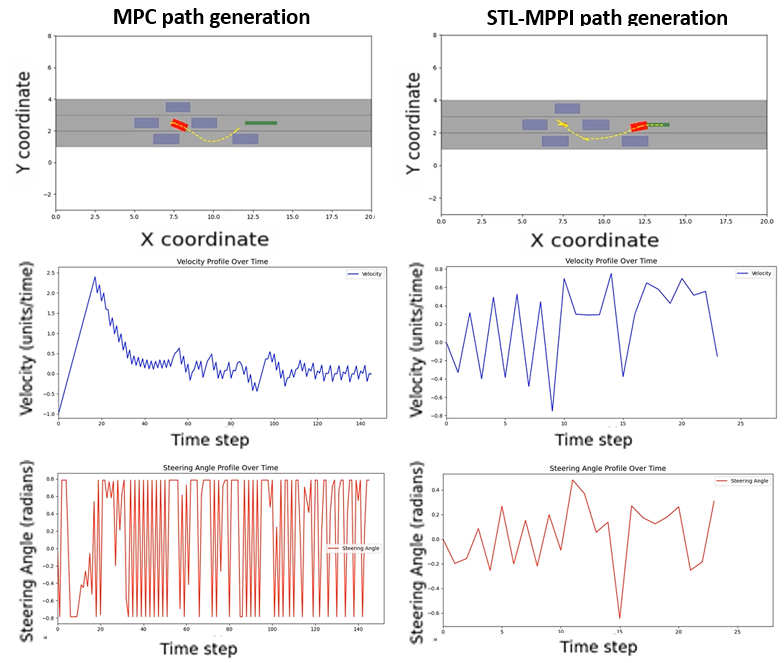}
        \caption{Comparison of the MPC method vs STL-MPPI for the lane deadlock scenario.
        The MPC method uses safety constraints as soft constraints to find feasible solutions, but produces suboptimal paths that get too close to obstacles. In contrast, the STL-MPPI approach generates a safer and more effective path operating back-forward maneuvers.}
        \label{fig:traffic1}
    \end{minipage}
        \begin{minipage}{\linewidth}
        \vspace{0.5cm}
    \includegraphics[width=\linewidth]{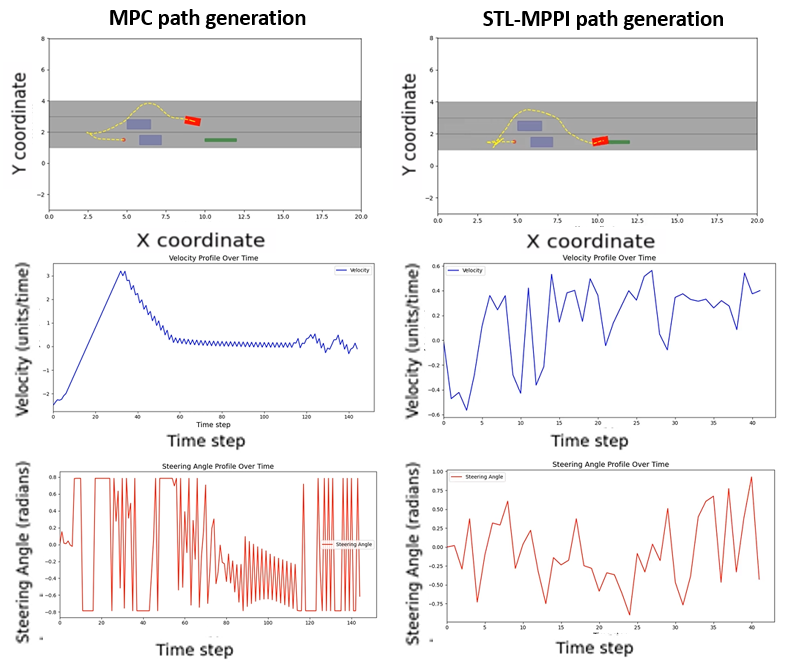}
        \caption{Comparison of the MPC method vs STL-MPPI for the traffic deadlock scenario.
        As before, the MPC method uses safety constraints as soft constraints to find feasible solutions going too close to obstacles. In contrast, the STL-MPPI approach generates a safer and more effective path to navigate around the obstacles.}
        \label{fig:traffic2}
    \end{minipage}
\end{figure}

\paragraph{Lane Deadlock}
Fig.~\ref{fig:traffic1} considers the case in which the ego vehicle reaches a stop due to an incident or road work that obstructs its original path. The ego vehicle needs to find a new path to overtake such an obstacle and proceed on its original path.  In Fig.~\ref{fig:traffic1}, we observe a comparative analysis of two path planning strategies: MPC and STL-MPPI control. The MPC path is shown to be suboptimal in this situation and it fails to navigate a new path effectively, resulting in a halt or a non-optimal rerouting. The STL-MPPI path demonstrates a more robust approach, successfully recalculating a path that navigates around the obstruction. The MPC finds an effective new path with significant delay (computational time: 1.7 seconds) compared to STL-MPPI's much quicker response (2.8e-5 seconds). 
This indicates that the STL-MPPI approach can handle unexpected situations more effectively than the standard MPC method.

\paragraph{Traffic Deadlock}
Fig.~\ref{fig:traffic2} examines a scenario where the 
AV is immobilized due to the presence of an obstacle on its planned path. The obstacle may be another vehicle or an unexpected stationary object. This scenario assumes that all vehicles in the vicinity of the AV are static, and that all road agents are cooperating, thereby providing the AV with the space required to perform its maneuver to disengage from the obstructed position.

 Fig.~\ref{fig:traffic2} shows that the MPC method demonstrates limited flexibility in its response, leading to either a premature stop or an inefficient rerouting of the AV with a computational time of 4 seconds, while the STL-MPPI navigates around the obstacle efficiently and quickly (4e-5 seconds), 
while maintaining compliance with traffic regulations, highlighting its potential for improved real-time navigational decisions in autonomous driving systems.


\subsection{Hardware Demonstration}
We extend validation studies by demonstrating STL-MPPI with a scaled RC car. In particular, we leverage the Multi-agent System for non-Holonomic Racing (MuSHR) \cite{mushr} autonomous vehicles at a test track of Honda Research Institute USA, Inc. in San Jose, CA. 

We use the Robot Operating System (ROS) to establish communications among sensors, actuators, and computing units. The MuSHR robot uses a LiDAR for determining its own state (position, velocity, and orientation) based on a given grid map of the track
and surrounding landmarks. The planner runs at 10Hz on Intel NUC mini PC onboard MuSHR. The testing scenarios are designed to evaluate the planner's ability to recover from a potential deadlock situation involving static obstacles. 

The goal of these tests is to assess the ego vehicle’s decision-making algorithms in conditions that require precise control and awareness of vehicle dimensions relative to the available space. It evaluates the vehicle’s ability to maintain safe passage through a narrow area that has the potential for deadlock if navigated improperly. 
Importantly, the tests are designed to include real-world complexities such as sensor noise, discrepancies between the theoretical models and actual vehicle behavior, and system latencies, all of which challenge the robustness of the proposed algorithms.

The static MuSHR cars are arranged to simulate common urban deadlock conditions, such as narrow lanes blocked by improperly parked vehicles or lanes closed for construction. The vehicles are placed such that there is no clear, direct path for the ego vehicle to proceed along its planned route without maneuvering around them. This setup is created to test the planner’s ability to recognize and resolve deadlock by finding a viable path that avoids the static obstacles.



\begin{figure}[h!]
    \centering
    \begin{minipage}{\linewidth}
        \centering
        \includegraphics[width=\linewidth]{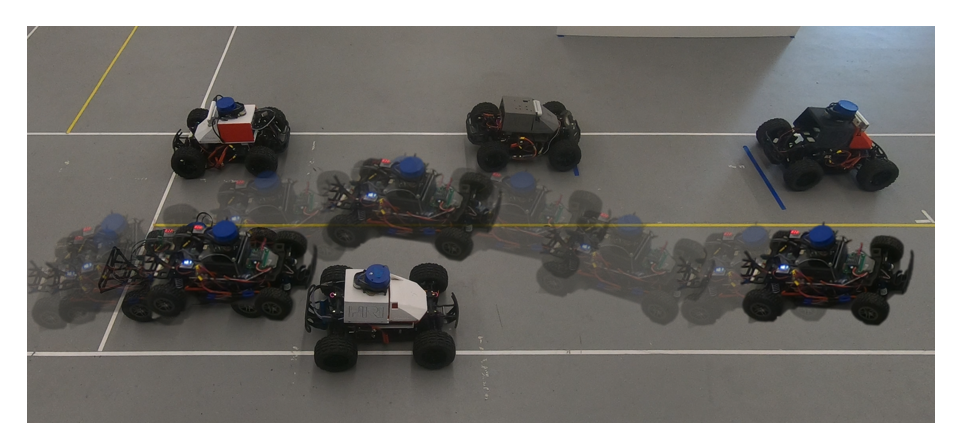}
        \caption{Demonstrating our algorithm on real hardware. The ego car needs to navigate around a car that stopped in front.}
        \label{case1}
    \end{minipage}
    \hfill  
    \begin{minipage}{\linewidth}
    \vspace{0.5cm}
        \centering
        \includegraphics[width=\linewidth]{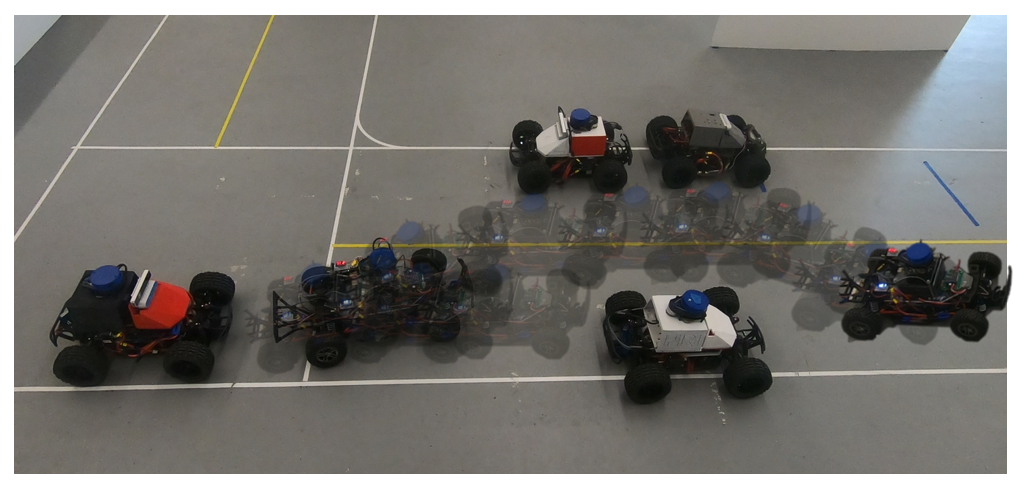}
        \caption{Demonstration where a second car is now positioned behind the ego car, limiting the space the ego car has available for backing up.}
        \label{case2}
    \end{minipage}
\end{figure}

 Figure~\ref{case1} illustrates a scenario where the MuSHR vehicles are arranged with one vehicle directly in front of the ego vehicle and others creating a narrow roadway in the other lane. This mimics a narrow street where the car in front of us stops unexpectedly to pick up a passenger. The ego vehicle is initially facing a direct obstruction, and there is no straightforward path to proceed without replanning its course.  

Fig.~\ref{case2} adds challenges to scenario by limiting space for reversing and passing the obstacles. This scenario represents situations like roadwork zones, narrowed lanes due to accidents, or streets where vehicles are parked along the sides, restricting the available driving space. The stationary vehicles serve as static obstacles, forcing the ego vehicle to maneuver carefully, as it would in a real traffic scenario where opposing vehicles are approaching in a confined space. 

Success in these scenarios is measured by the ego vehicle’s ability to maintain a steady speed, accuracy on its trajectory, maintaining minimal distance from the stationary vehicles without collisions, and the time taken to clear the passage. 
In both these testing scenarios, STL-MPPI has shown effectiveness and potential for practical implementation in real-world scenarios. 

\begin{remark}
We have not observed any collisions by carefully choosing the penalty weights in the MPPI framework. However, it's important to note that the MPPI system inherently employs soft constraints, which means that it does not guarantee collision avoidance in every scenario. To improve safety measures, one effective strategy could be to adaptively adjust the penalties based on the assessed risks. 
\end{remark}




\section{Conclusion and Limitations}

We proposed a method for deadlock recovery that combines Hybrid A$^\star$, STL and MPPI. 
Through the integration of STL within an MPPI controller, we combine the robustness of temporal logic with the adaptability of MPPI, enabling more responsive and flexible path planning. 
The use of STL within the MPPI controller allows for the incorporation of high-level, time-bound specifications into the vehicle's operational framework, ensuring adherence to safety constraints and road regulations without reducing computational speed. This contrasts with MPC-based systems, where integrating such logical constraints often results in increased computational demands. 
This combination has been validated in real-world hardware applications, confirming its practical benefits and effectiveness in on-the-ground vehicle operation. 

 Future works include scaling up in more complex environments with a higher number of agents or more complicated interactions with dynamical agents to ensure that it can operate with the same efficacy in scenarios with more unpredictable environmental factors. 
 Furthermore, navigating environments with dynamic agents, such as other vehicles, requires predictive models to anticipate the reactions of these agents to the autonomous vehicle’s maneuvers. 

\bibliographystyle{unsrt}
\bibliography{refs.bib}


\end{document}